# A comprehensive review on Plant Leaf Disease detection using Deep learning

Sumaya Mustofa
Department of Computer Science and Engineering
Daffodil International University, Dhaka, Bangladesh
sumaya15-3445@diu.edu.bd

Md Mehedi Hasan Munna
Department of Computer Science and Engineering
Daffodil International University, Dhaka, Bangladesh
mehedi15-3037@diu.edu.bd

Yousuf Rayhan Emon
Department of Computer Science and Engineering
Daffodil International University, Dhaka, Bangladesh
yousuf15-3220@diu.edu.bd

Golam Rabbany
Lecturer
Department of CSE
Daffodil International Univarsity
Dhaka, Bangladesh
rabbany.cse@diu.edu.bd

Dr. Md Taimur Ahad
Associate Professor
Department of CSE
Daffodil International Univarsity
Dhaka, Bangladesh
taimur.cse0396.c@diu.edu.bd

*Abstract–* *Leaf disease is a common fatal disease for plants. Early diagnosis and detection is necessary in order to improve the prognosis of leaf diseases affecting plant. For predicting leaf disease, several automated systems have already been developed using different plant pathology imaging modalities. This paper provides a systematic review of the literature on leaf disease-based models for the diagnosis of various plant leaf diseases via deep learning. The advantages and limitations of different deep learning models including Vision Transformer (ViT), Deep convolutional neural network (DCNN), Convolutional neural network (CNN), Residual Skip Network-based Super-Resolution for Leaf Disease Detection (RSNSR-LDD), Disease Detection Network (DDN), and YOLO (You only look once) are described in this review. The review also shows that the studies related to leaf disease detection applied different deep learning models to a number of publicly available datasets. For comparing the performance of the models, different metrics such as accuracy, precision, recall, etc. were used in the existing studies.*

# Introduction

The success of transformer-based models, such as Vision Transformer (ViT), in language processing motivated data scientists to unitize ViT for classification, detection, and localization of plant disease in vision-based deep learning. The usage of vision transformers (ViT), which have proven to be effective models applies self-attention mechanism band transformers to vision problems (Carion et al. 2020; Chen et al. 2018; Ramachandran et al. 2019; Vaswani et al. 2021). Due to recent developments in various fields, ViT has gained popularity, making it an excellent option for image processing in the agricultural industry as well. However, as stated by Deshpande et al. (2022), despite the fact that numerous computer vision and artificial intelligence-based schemes have been put forth in the past for automatic plant leaf disease detection, their performance has been found to be insufficient due to poor feature representation, lower-order correlation of raw features, data imbalance issues, and a lack of generalization. The trusted answer to this issue is Vision Transformer. Self-supervised ViT features offer explicit information on the semantic segmentation of an image, which does not appear as clearly with supervised ViTs or with CNN, according to a study by Caron et al. (2021)

The Vision Transformer (ViT) structure, based on how individuals classify images of specific elements, was recently introduced to help segmentation applications. When a person looks at a photograph, they focus in a certain part of the image to discover the object of interest, according to Borhani et al. (2022). This methodology is applied by the ViT structure for picture categorization. Vision transformer (ViT) with hard patch embedding as input is suggested by Dosovitskiy et al. (2020). In order to encode the spatial location of each patch within the image, ViT also uses positional embeddings.

Recent years have seen an increase in leaf diseases due to climate change, the expansion of outdoor air pollution, and global warming. It has a big effect on how productive agriculture is. Farmers used to make intuitive diagnoses of leaf diseases, but this method is unreliable and inefficient. With the advancement of deep learning, many CNN models have pioneered their way to the identification of plant-leaf illnesses to cut down on farmers' work. However, these models are only capable of detecting particular crops and not so efficient during diseases. Plant leaf disease has the potential to significantly reduce the number of agricultural products produced on each farm, but Vision Transformer can give farmers visual information so that they can take the required precautions. By taking significant elements from the leaf image, ViT can pinpoint the precise area of the leaf where the disease is present, giving farmer's useful information. According to Chougui et al. (2022), ViT can quickly classify and identify various plant-leaf diseases with high accuracy results. ViT can also achieve excellent classification performance by automatically extracting those necessary elements for classification from photos. A crucial factor in the effectiveness of vision transformers is image quality. However, when the input image resolution is poor, current plant leaf disease identification algorithms do not give sufficient disease detection accuracies, despite the fact that new deep learning approaches have significantly aided in the detection of plant leaf diseases. Solutions for smart agriculture are developing that integrate deep learning and computer vision for the early diagnosis and management of diseases in order to address this problem. For the real-time detection and diagnosis of leaf diseases, these systems employ deep learning techniques based on vision. In this way to broaden its use, Vision Transformer is combining other algorithms. However, vision transformers have hardly ever been researched for use in agricultural pathology.

Bangladesh is renowned for being an agriculturally oriented nation, and like many other emerging nations, it has historically seen agriculture as its primary industry. The agriculture sector has a considerable impact on Bangladesh's gross domestic product (GDP). But agricultural production might significantly decrease, which would have a negative impact on Bangladesh as well as the world's food security due to plant disease especially plant leaf disease. Crop productivity is in danger due to the ongoing spread of leaf diseases. The presence of diseases on the leaves, which can have a negative impact on a plant's lifecycle. One of the major issues with smart agriculture is plant-leaf disease was addressed by Rethik et al. (2023) associated with cultivating plants is the prevalence of diseases on the leaves, which can significantly impact the plant's growth and development. Plant-leaf disease is one of the key problems of smart agriculture which has a significant impact on the global economy. Each year, the Food and Agriculture Organization of the United Nations (FAO) estimates that plant pests cause the loss of 10–16% of the annual global harvest, or $220 billion USD. Due to the diversity of plant species and the regional features of many of these species, studies in this area have not been undertaken to the desired level. It is apparent that early identification of illnesses on plant leaves remains a challenging challenge, even though many researchers have investigated diseases on plant leaves. The illnesses, which are brought on by a range of infections and environmental stressors, can take many different forms. The signs of infection on infected leaves can differ from one plant to another or even from one leaf to another on the same plant. It has been challenging to develop an efficient plant leaf disease detection algorithm because, according to the "International Society for Plant Pathology," plants are vulnerable to a variety of 137 pathogens and pests, including bacteria, nematodes, viruses, and over 19,000 fungi (Jain et al. 2019) are known to cause diseases in crop plants leaves worldwide. Plant leaf diseases are a serious worry for farmers all over the world because they can cause crops to suffer severe financial losses.

The aim of this literature review study is to connect past, present, and future research in plant leaf disease detection using transformer-based CNN. Firstly, we provide previous and current studies focusing on plant leaf disease detection using ViT. We also aimed to know how the vision transformer (ViT) gradually improved the accuracy. Secondly, we wanted to know about research studies conducted in various agricultural countries on plant leaf diseases. Furthermore, we wanted to understand future research directions and the deep learning algorithms evolving, especially ViT, through the years in agricultural research.

# 1. Process of Literature Review

Studies related to transformer based CNN for detection, feature extraction, feature extraction using traditional machine learning techniques for the classification, auto encoders (AEs), Vision transformer, hybrid model for deep learning techniques included in this study. Most importantly, the research that experimented using a proper research methodology without providing experimental research were included in the literature review.

**Article Identification**

The literature review of this study followed Preferred Reporting Items for Systematic Reviews and Meta-Analyses (PRISMA) guidelines. A review of around thirty-five selected papers is presented in this paper. All of the articles cover the classification and detection of plant leaf using transformer-based deep learning techniques. Figure 1 discussed the process of article selection is described:

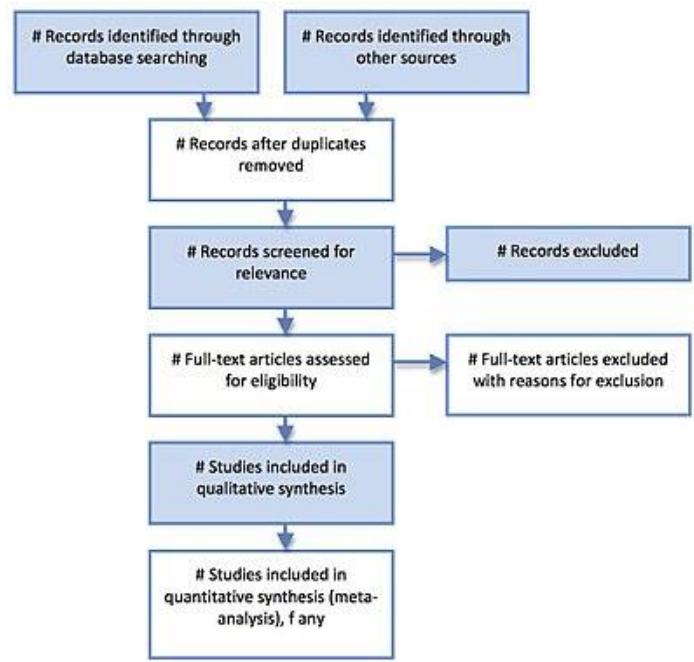

Figure 1: The process of article selection

**Article Selection**

Articles were selected for final review using a three-stage screening process based on a series of inclusion and exclusion criteria. After removing duplicate records that were generated from using two databases, articles were first screened based on the title alone. The abstract was then assessed, and finally, the full articles were checked to confirm eligibility. The entire screening process was conducted by the chief investigator.

The meet the inclusion criteria, articles had to:

- Be concerned with the application of transformer based deep learning techniques for plant leaf disease classification.
- Included articles were limited to those published from 2018 to 2023 to focus on deep learning methodologies. Here, a study was defined as work that employed a transformer-based deep learning algorithm to classify, detect plant leaf disease and that involved the use of one or more of the following performance metrics: accuracy, the area under the receiver operating characteristics curve, sensitivity, specificity, or F1 score.

Exclusion criteria were:

- Preprint studies

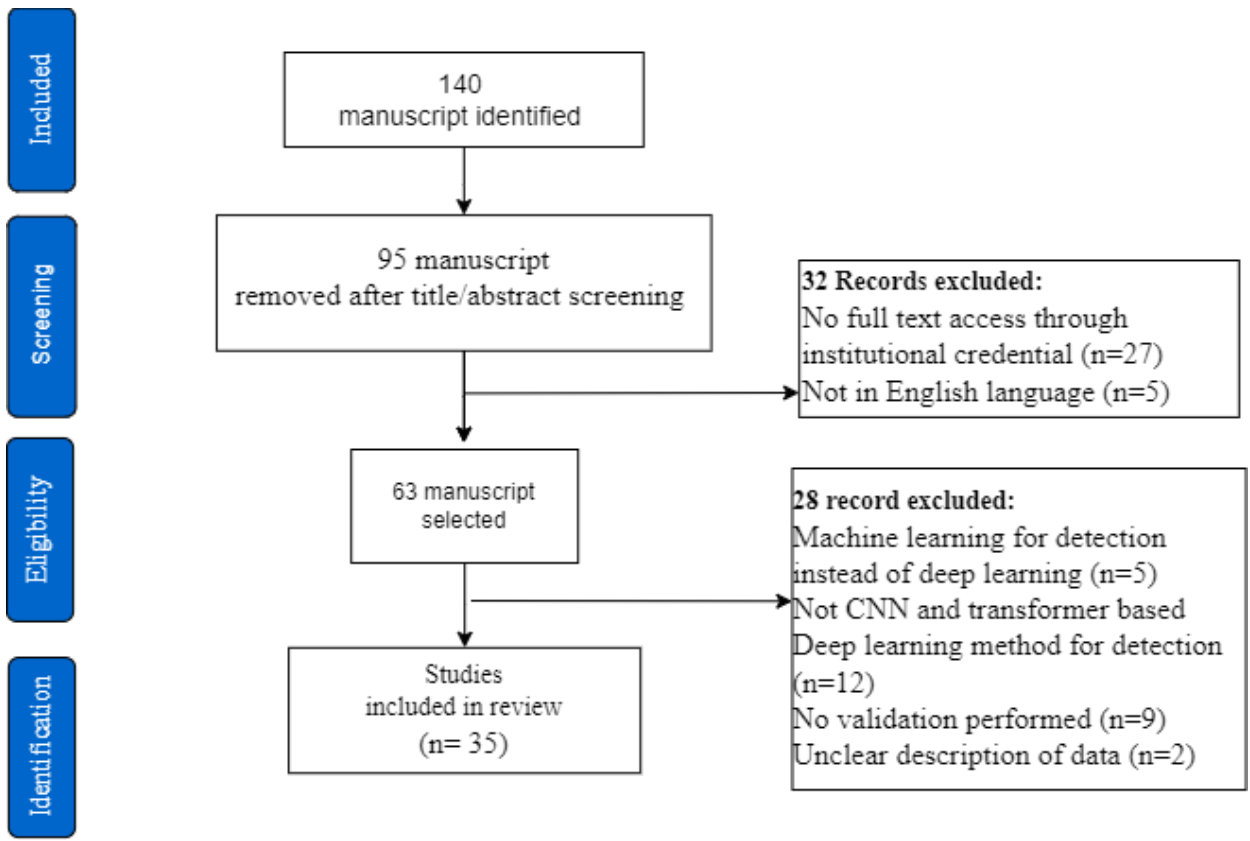

Figure 2: The Preprint studies.

## 2. Literature review

Borhani et al. (2022) developed a deep learning-based method for automated plant disease categorization utilizing a vision transformer in order to give farmers visual information. According to the scientists, the real-time automated plant disease classification method is built on Vision Transformer (ViT), making the deep learning technique very lightweight. For the categorization of plant diseases, conventional convolutional neural network (CNN) techniques and CNN + ViT combos have also been used in addition to the ViT. To speed up prediction, the model coupled CNN blocks with attention blocks. Authors' approached model 3 and 4 for the corresponding wheat rot, rice leaf, and plant village datasets exhibit the maximum convergence accuracy. To generate a higher accuracy, the RGB version of the photos has been employed. But correctness of the model was missing in the study and convergence score also isn't an ideal metric in deep learning based experiment.

Bandi et al. (2023) proposed a model for plant leaf disease stage categorization and detection that operates according to the severity of leaf infection. You only look once version 5 (YOLOv5) deep learning model is used to detect plant leaf disease. The background of the diseased leaf is then removed using U2-Net architecture, and stage classification is then carried out using a vision transformer (ViT) to categorize. The apple leaf is the major focus of this work when executing stage categorization. With a confidence level of 0.2, the YOLO v5 can obtain a maximum f1-score of 0.57, whereas the vision transformer can reach a f1-score of 0.908 with or without a backdrop image.

Rethik et al. (2023) proposed attention-based mapping for plant leaves using Vision Transformer to categorize illnesses. In this study, researchers used Vision Transformer instead of CNN to categorize plant leaf diseases. The test accuracy attained by the three vision transformer models under comparison in this is 85.87%, 89.16%, and 94.16%, respectively. The models are ViT1, ViT2, and pre-trained ViT_b16. The findings demonstrate that the suggested model is capable of pinpointing the precise area of the leaf where the illness is present, giving farmers useful information.

Chougu et al. (2022) described plant-leaf diseases classification using CNN, CBAM and Vision Transformer. The authors developed four pretrained models using huge datasets like MobileNet, VGG-16, VGG-19, and ResNET, and suggested a deep convolutional neural network architecture with and without attention mechanisms. Additionally, authors' adjusted two ViT models: the Vit B32 from Keras and the Google base patch 16. The suggested model achieved a 97.74% accuracy rate. The accuracy of the pre-trained models was up to 99.52%. And the ViT models achieved up to 99.7% accuracy.

Sharma et al. (2023) presented a new deeper lightweight convolutional neural network architecture (DLMC-Net) to perform plant leaf disease detection across multiple crops for real-time agricultural applications. The passage layer and a series of collective blocks are added in the suggested model in order to extract deep characteristics. These advantages include feature reuse and propagation, which solve the vanishing gradient issue. Convolution blocks that are point-wise and separable are also used to lower the number of trainable parameters. On eight metrics, including accuracy, error, precision, recall, sensitivity, specificity, F1-score, and Matthews correlation coefficient, experimental results of the proposed model are compared against seven state-of-the-art models. Even with complex background conditions, the suggested model outperformed all other models, with accuracy values of 93.56%, 92.34%, 99.50%, and 96.56% on the datasets for citrus, cucumber, grapes, and tomatoes, respectively.

Deshpande et al. (2022) addressed automatic plant leaf disease detection using Deep Convolutional Neural Network (DCNN) to increase the feature representation and correlation and Generative Adversarial Network (GAN) for data augmentation to cope up with data imbalance problem. Based on accuracy, precision, recall, and F1 score, which have significantly improved over conventional methods for the plant leaf disease database (accuracy of 99.74%, precision, recall, and F1 score of -0.99), the success of the suggested strategy is assessed.

Hossain et al. (2023) addressed a study to analyze the effects of transformer-based approaches that aggregate different scales of attention on variants of features for the classification of tomato leaf diseases from image data. Four state-of-the-art transformer-based models, namely, External Attention Transformer (EANet), Multi-Axis Vision Transformer (MaxViT), Compact Convolutional Transformers (CCT), and Pyramid Vision Transformer (PVT), are trained and tested on a multiclass tomato disease dataset. The result analysis showcases that MaxViT comfortably outperforms the other three transformer models with 97% overall accuracy, as opposed to the 89% accuracy achieved by EANet, 91% by CCT, and 93% by PVT. MaxViT architecture is the most effective transformer model to classify tomato leaf disease because it achieves a smoother learning curve compared to the other transformers.

Thai et al. (2021) developed a novel model named Vision Transformer (ViT) in place of a convolution neural network (CNN) for classifying cassava leaf diseases. On the dataset for cassava leaf disease, experimental results demonstrate that this model can achieve competitive accuracy that is at least 1% greater than well-known CNN models likr EfficientNet, Resnet50d . This research also demonstrated that the ViT model is successfully implemented into the Raspberry Pi

4, an edge device that can be connected to a drone to enable farmers to quickly and effectively find diseased leaves.

Li et al. (2022) proposed an automatic pest identification method based on the Vision Transformer (ViT). The plant diseases and insect pests data sets are improved using techniques including Histogram Equalization, Laplacian, Gamma Transformation, CLAHE, Retinex-SSR, and Retinex-MSR in order to prevent training overfitting. According to the simulation results, the built-in ViT network has a test recognition accuracy rate of 96.71% on the publicly available Plant_Village dataset of plant diseases and insect pests, which is about 1.00% higher than the method for identifying plant diseases and pests based on conventional convolutional neural networks like GoogleNet and EfficentNetV2.

Yeswanth et al. (2023) proposed a novel Residual Skip Network-based Super-Resolution for Leaf Disease Detection (RSNSR-LDD) in the Grape plant. The super-resolution (SR) image is produced using a decoding block and a convolutional layer. For training, a brand-new collaborative loss function is suggested. The Disease Detection Network (DDN) receives the acquired SR picture in order to identify grape leaf disease. With numerous super-resolution scaling factors for different grape leaf pictures, the proposed model was thoroughly trained and evaluated on the PlantVillage, Grape 400, and Grape Leaf Disease datasets. The proposed model RSNSR-LDD attained accuracies of 97.19%, 99.37%, and 99.06% for the PlantVillage dataset, 96.88%, 97.12%, and 95.43% for the Grape400 dataset, and 100% for the Grape Leaf Disease dataset for various super-resolution scaling factors like X2, X4, and X6.

Thai et al. (2023) developed an efficient vision transformer for Cassava Leaf Disease detection. The model FormerLeaf, a transformer-based model for detecting leaf disease, and two strategies for improving the model's performance. To choose the most crucial attention heads for each layer in the Transformer model, the authors suggested the Least Important Attention Pruning (LeIAP) algorithm. It might cut the size of the model by up to 28%, speed up evaluation by 15%, and improve accuracy by roughly 3%. In order to determine matrix correlation in the model, it also used sparse matrix-matrix multiplication (SPMM). Due to the model's complexity being reduced from $O(n^2)$ to $O(n^2/p)$, training time is cut by 10% but performance is kept the same.

Alshammari at al. (2022) developed a unique deep ensemble learning strategy that combines the convolutional neural network model with vision transformer model. This approach aims to identify and categorize diseases that may impact olive leaves. Olive leaf disease was categorized using deep convolutional models-based binary and multi classification systems. The outcomes are encouraging and demonstrate the potency of combining CNN and vision transformer models. With an accuracy of roughly 96% for multiclass classification and 97% for binary classification, the model outperformed the competition.

Zhou et al. (2023) proposed a residual-distilled transformer architecture in this study for feature extraction and prediction, a multi-layer perceptron (MLP) is fed with the residual concatenation of the vision transformer and the distillation transformer. On the dataset for rice leaf disease collected in paddy fields, experimental results show that the proposed method outperforms the current state-of-the-art models and obtains a 0.89 F1 score and 0.92 top-1 accuracy.

Li et al. (2023) presented Shuffle-convolution-based lightweight Vision transformer for effective diagnosis of sugarcane leaf diseases named SLViT . The SLViT hybrid network is initially trained on the freely available disease dataset Plant Village and the independently created sugarcane leaf disease dataset SLD10k. The transformer encoder is converted to a flexible plug-in (LViT) and then integrated into several locations of a lightweight CNN architecture (SHDC). The experiments show that all of SLViT's modifications have improved the system's performance as a whole. On Plant Village, SLViT outperforms three specially created leaf-disease recognition models and six SOTA models in terms of speed (1,832 FPS), weight (2 MB), consumption (50 M), and precision (98.84%). On the SLD10k dataset, SLViT outperformed MobileNetV3_small with an accuracy boost of 1.87% and a size reduction of 66.3%.

Thakur et al. (2022) developed a Vision Transformer enabled Convolutional Neural Network model called "PlantXViT" for plant disease identification. The suggested model effectively recognizes a wide variety of plant illnesses for various crops by combining the abilities of conventional convolutional neural networks with the Vision Transformers. The suggested model is appropriate for IoT-based smart agriculture services since it has a lightweight structure and only 0.8 million trainable parameters. On all five datasets, the proposed PlantXViT network outperforms five cutting-edge techniques. Even with difficult background conditions, the average accuracy for identifying plant diseases is shown to exceed 93.55%, 92.59%, and 98.33% on the datasets for apples, maize, and rice, respectively. Gradient-weighted class activation maps and Local Interpretable Model Agnostic Explanation are used to assess the effectiveness of the given model's explainability.

Li et al. (2021) proposed the RegNet novel lightweight convolutional neural network which was used to detect Apple Leaf illness using a tiny and unbalanced dataset,. A number of comparative experiments using cutting-edge convolutional neural networks (CNNs), including ShuffleNet, EfficientNet-B0, MobileNetV3, and Vision Transformer, were carried out to evaluate the efficacy of the RegNet model. With a learning rate of 0.0001, RegNet-Adam achieved an overall accuracy of 99.23% on the test set and an average accuracy of 99.8% on the validation set.

Mahbub et al. (2023) proposed a lightweight convolutional Neural Network (LCNN) to detect mango leaf disease. Authors' classified seven mango leaf disease and healthy leaf and applied pre-trained model named VGG16, Resnet50, Resnet101, and Xception. The accuracy of LCNN model was highest with the accuracy performance of 98% which was better than other pre-trained model.

Mehta et al. (2023) proposed a new method for identifying and categorizing mango leaf illnesses using a Convolutional Neural Network (CNN) model based on federated learning. The suggested model on four distinct customers while concentrating on the five disease classifications Healthy, Anthracnose, Powdery Mildew, Leaf Spot, and Leaf Curl. The experiment result shows, Precision values ranging from 93.33% to 96.01%, recall values ranging from 90.59% to 97.45%, F1-scores ranging from 92.64% to 96.10%, and accuracy values between 97% and 98%. The macro, weighted, and micro averages, with macro averages ranging from 93.18% to 94.97%, weighted averages ranging from 93.26% to 95.08%, and micro averages ranging from 93.26% to 95.08%. By employing federated learning to secure data privacy, this methodology enables customers to collaborate and benefit from shared learning without putting their data at risk.

Zhuang et al. (2021) proposed a deep learning method to identify disease of cassava leaf image based on vision transformer. The experiment results show that Vision-Transformer-based model can effectively achieve an excellent performance after applying the K-Fold cross validation technique. The model reached a categorization accuracy 0.9002 on the private test set regarding the classification of cassava leaf diseases.

Zeng et al. (2022) proposed an image classification model for large-scale and fine-grained diseases named Squeeze-and-Excitation Vision Transformer (SEViT) to solve difficulties to distinguish similar diseases, which does not perform well in large-scale and fine-grained disease diagnosis tasks. SEViT includes ResNet embedded with channel attention module as the preprocessing network, ViT as the feature classification network. The experimental results show that the classification accuracy of SEViT in the test set achieves 88.34%. Compared with the baseline model, the classification accuracy of SEViT is improved by 5.15%.

Table 1: Research Matrix.

| Author | Model | Dataset | Accuracy | Contribution |
|---|---|---|---|---|
| Ahad et al. (2023) | Ensemble and transfer learning | Rice leaf | 98% | Provided a new ensemble model. |
| Borhani et al. (2022) | Combination (CNN,ViT) | Wheat Rust Classification Dataset (WRCD), Rice Leaf Disease Dataset (RLDD), Plant village. | - | Proposed a lightweight deep learning approach combining CNN and ViT. |
| Bandi et al. (2023) | you only look once version5 (YOLOv5), U2-Net, ViT | PlantDoc, PlantVillage | 90.8 % | Stage classification techniques on Apple leaf disease. |
| Rethik et al. (2023) | ViT1, ViT2, ViT_b16 | - | 85.87%(ViT1), 89.16%(ViT2), 94.16%(ViT_b16) | Identifying specific affected region of leaf. |
| Chougu et al. (2022) | CNN(without attention mechanism), MobileNet, VGG-16, VGG-19, ResNET, vit b32, base patch 16 | PlantVillage, Tomato | 97.74% (pretrained), 99.7%(ViT) | Proposed a deep CNN without attention mechanism. |

| Sharma et al. (2023) | DLMC-Net | Citrus, Cucumber, Grapes, and Tomato | 93.56%(Citrus), 92.34%(Cucumber), 99.50%(Grapes), 96.56% (Tomato) | A deeper lightweight multi-class classification model. |
|---|---|---|---|---|
| Deshpande et al. (2022) | Deep Convolutional Neural Network (DCNN), Generative Adversarial Network (GAN) | Tomato plant disease, Plant Village leaf disease | 99.74%, | Model to increase the feature representation and correlation, solution to data imbalance problem. |
| Hossain et al. (2023) | External Attention Transformer (EANet), Multi-Axis Vision Transformer (MaxViT), Compact Convolutional Transformers (CCT), Pyramid Vision Transformer (PVT) | Multi class tomato diseases | 97% (MaxViT) | Aggregating different scales of attention on feature variants. |
| Thai et al. (2021) | Vision Transformer | Cassava Leaf Disease. | 1% higher than CNN. | Deploying the model in Raspberry Pi 4 device, that can attached to a drone to automatically detect infected leaves. |
| Li et al. (2022) | Vision Transformer | Plant Village | 96.71% | Higher accuracy than traditional CNN. |
| Yeswanth et al. (2023) | Residual Skip Network-based Super-Resolution for Leaf Disease Detection (RSNSR-LDD), | PlantVillage, Grape 400, Grape Leaf Disease | 97.19%, 99.37%, 99.06% (PlantVillage), 96.88%, 97.12%, 95.43% (Grape400), 100% | Proposed a novel Residual Skip Network-based Super-Resolution for Leaf Disease Detection (RSNSR-LDD) |

|  | Disease Detection Network (DDN) |  | (Grape Leaf Disease) |  |
|---|---|---|---|---|
| Thai et al. (2023) | Least Important Attention Pruning (LeIAP), Vision Transformer | Cassava Leaf Disease. | 3% accuracy enhancement. | Reducing model size, Accelerate evaluation speed, Accuracy enhancement, Complexity Reduction. |
| Alshammari at al. (2022) | Ensemble (ViT+CNN) | Olive Disease. | 96% (multiclass classification), 97% (binary classification) | Proposed ensemble vision transformer with CNN. |
| Zhou et al. (2023) | MLP, ViT, distilled transformer. | Rice Leaf disease | 92% | Proposed a a residual-distilled transformer architecture. |
| Li et al. (2023) | SLViT | Plant Village, Sugarcane leaf disease dataset SLD10k | Accuracy Enhancement 1.87%. | Proposed a Shuffle-convolution-based lightweight Vision transformer. |
| Thakur et al. (2022) | PlantXViT | Apple, Maize, Rice | 93.55%(Apple), 92.59% (Maize), 98.33% (Rice) | Proposed a ViT enabled CNN model called "PlantXViT". |
| Li et al. (2021) | RegNet | Apple leaf disease | 99.8% (validation), 99.23% (test) | Proposed a new lightweight convolutional neural network RegNet. |
| Mahbub et al. (2023) | LCNN, VGG16, Resnet50, Resnet101, Xception. | - | 98% | Proposed a LCNN model. |
| Mehta et al. (2023) | CNN (federated based learning) | - | Between (97%-98%) | A federated learning-based CNN that solves the challenges farmers face in recognizing and controlling mango leaf diseases has been proposed. This will lead to more productive |

| | | | | and long-lasting agricultural practices. |
|---|---|---|---|---|
| Zhuang et al. (2021) | Vision Transformer | Cassava leaf diseases | 90% | Enhanced performance of vision transformer applying k-fold cross validation. |
| Zeng et al. (2022) | SEViT (embedded) | - | 88.34% | Proposed a model to solve difficulties to distinguish similar diseases, which does not perform well in large-scale and fine-grained disease diagnosis tasks |
| Wang et al. (2022) | Swin Transformer | Cucumber leaf disease. | STA-GAN: 98.97%, 96.81%, 94.85% and 90.01% SwinT, original SwinT, EfficientNet-B5 and ResNet-101: increasing by 2.17%, 3.62%, 2.13% and 11.23% | Used improved Swin Transformer to eradicate challenge of small sample size insufficient data and complex background. |
| Salamai et al. (2023) | Visual Transformer | Paddy disease. | 98.74% | Introduced a lesion-aware visual transformer for accurate and reliable detection of paddy leaf diseases through identifying discriminatory lesion features. |

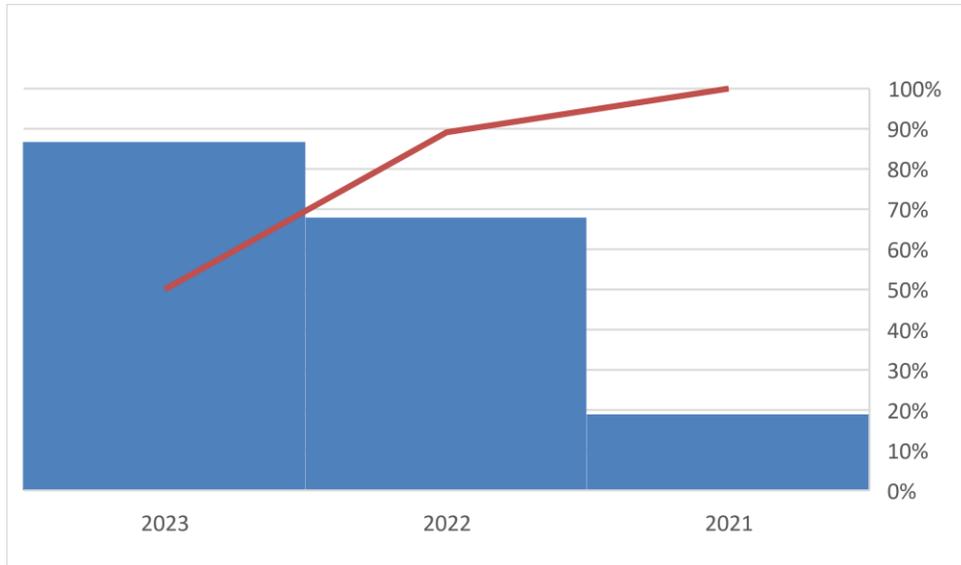

Figure 3: Accuracy changes through the year of literature review.

## 3. Inference from current research studies

Despite the growing interest in deep learning and computer vision, the research points out that there is a lack of investigation into the application of Vision Transformers specifically within the field of agricultural pathology. This suggests a potential research gap that the paper may address. Deep learning techniques are examined in this study to detect plant leaf diseases. Inference from current research studies, we can conclude the following points.

1. The research matrix (table 1) offers a number of significant directions for the current study. According to the literature review, deep learning provides superior outcomes. This research work makes use of multiple enhancements for enhanced performance. Combination of CNN and ViT was a good approach compared with the other hybrid techniques in [1]. It concluded from [2] that you only look once version5 (YOLOv5), U2-Net, ViT performed better in apple leaf disease. From [3], it showed that ViT1, ViT2, ViT_b16 give the better results on specific affected area of leaf. It gave better results on tomato disease. From [4], the author used CNN (without attention mechanism), MobileNet, VGG-16, VGG-19 , ResNET, vit b32, base patch 16. In paper [5], the author proposed a deeper lightweight multi-class classification model

(DLMC-Net) on Citrus, Cucumber, Grapes, and Tomato plant disease in terms of performance. From [6], the author used Deep Convolutional Neural Network (DCNN), Generative Adversarial Network (GAN) techniques and found the better result for tomato plant disease. It achieved the 99.74% accuracy. The author discussed External Attention Transformer (EANet), Multi-Axis Vision Transformer (MaxViT), Compact Convolutional Transformers (CCT), Pyramid Vision Transformer (PVT) on multiclass tomato disease and found the 97% better results as compared to other methodologies in [7]. In paper [8], the research work was conducted on Cassava Leaf Disease using Vision Transformer with 1% higher than CNN and achieved better result. On the other hand from [11] we get the 3% higher performance on Cassava Leaf Disease using Least Important Attention Pruning (LeIAP) and Vision Transformer. In the paper [10], the author proposed the model Residual Skip Network-based Super-Resolution for Leaf Disease Detection (RSNSR-LDD), Disease Detection Network (DDN) and found the better result for the Grape Leaf Disease. In the paper [12], the author proposed the model Ensemble (ViT + CNN) and compared it with the other deep learning techniques and achieved better performance through CNN. In the paper [14], the author worked on Accuracy Enhancement 1.87% of Plant Village, Sugarcane leaf disease dataset SLD10k proposing model is SLViT and achieved the better performance. Proposed a ViT enabled CNN model called "PlantXViT" in [15] on Apple, Maize, and Rice leaf diseases getting the better performance. Author used in [33] improved Swin Transformer to eradicate challenge of small sample size insufficient data and complex background on Cucumber leaf disease for increasing the performance percentage.

2. This paper focus on the exploration of using deep learning techniques, specifically Vision Transformer (ViT), in language processing motivated data scientists to unitize ViT for classification, detection and management of various vegetables, fruits, flowers and cereals plant leaf diseases in the context of smart agriculture.

# 3. Limitations

The paper seeks to comprehensively review the current plant leaf disease detection using deep learning techniques, while also highlighting the underexplored area of applying Vision Transformers to agricultural pathology. This endeavor aligns with the broader goal of leveraging technology to enhance disease management practices and overall agricultural productivity. But In the literature, most of the models used publicly available free datasets for various vegetables and cereals plant leaf diseases detection. This reason the deep learning methods used for disease classification actually require a large number of annotated images, which is difficult to obtain. A large dataset of plant leaf disease images requires many images with accurately labeled instances of various diseases, and this process necessitates the expertise of individuals such as botanists or agricultural specialists. So we see the following limitations in this study –

1. The creation of such datasets is thus inherently time-consuming and challenging. Additionally, the diversity of plant diseases, variations in their appearance due to factors like lighting and growth stages, and the need for precise annotations further compound the complexity of dataset generation. This limitation becomes a bottleneck in training deep learning models effectively, as their performance heavily relies on the quality and diversity of the training data. Consequently, researchers and practitioners in the field of agricultural pathology often face the dual challenge of acquiring a sufficiently large and diverse dataset of labeled images while also ensuring the accuracy of annotations, which requires the involvement of domain experts and significant time investment. Hence, researchers prefer to use existing publicly available datasets.

2. This paper shows the different models which applied to public datasets with maximum accuracy but it may not be the optimum solution. According to the literature review, Vision Transformer improve the 1% accuracy than CNN. On the other hand for getting 3% higher performance, Least Important Attention Pruning (LeIAP) and Vision Transformer are used for detection. Some studies preferred Ensemble (ViT + CNN) and compared it with the other deep learning techniques and achieved better performance through CNN. However, CNN does not work well for high dimensional data. On the other hand, External Attention Transformer (EANet), Multi-Axis Vision Transformer (MaxViT), Compact Convolutional Transformers

(CCT), Pyramid Vision Transformer (PVT), SLViT and a ViT enabled CNN model called "PlantXViT" work well for high dimensional multicast image datasets.

## 4. Direction for Future Research

Emphasizes the integration of deep learning methodologies into the field of agriculture. By utilizing advanced deep learning techniques, the goal is to enhance the accuracy and efficiency of disease detection processes. Methodologies process the images using a transformer architecture with a self-attention mechanism, and their findings in image classification, object identification, and image segmentation have been promising. This section represents novel research directions aimed at advancing the classification of plant leaf diseases. One promising direction for improving the classification of these diseases is the application of reinforcement learning methods to support the diagnostic process. In this context, the architecture of the system and the fine-tuning of its parameters emerge as significant challenges that require exploration.

Additionally, a cutting-edge avenue for future research involves the integration of hybrid deep learning techniques. This hybrid approach holds the promise of leveraging the strengths of different techniques to achieve more comprehensive and reliable results. Another innovative approach to plant disease classification is the utilization of case-based reasoning. This method involves solving novel classification problems by drawing from solutions to similar problems encountered in the past. In summary, the future of research in the field of plant disease classification holds exciting prospects, including the adoption of reinforcement learning to push the boundaries of disease classification accuracy, adaptability, and efficiency, contributing to more effective disease management strategies in the context of smart agriculture.

## 5. Conclusion

This study represents a review for the emergence of smart agricultural solutions that incorporate in computer vision, vision transformers (ViT) are a relatively new and intriguing breakthrough. ViT can quickly classify and identify various plant-leaf diseases with high accuracy results and researchers have focused the strengths and weaknesses of some image classification and object identification models such as Vision Transformer (ViT), Deep convolutional neural network (DCNN), Convolutional neural network (CNN), Residual Skip Network-based Super-Resolution for Leaf Disease Detection (RSNSR-LDD), Disease Detection Network (DDN), and YOLO(You only look once) Moreover, after the basic concept of Vit, there have also Least Important Attention Pruning (LeIAP), Ensemble (ViT + CNN), External Attention Transformer (EANet), Multi-Axis Vision Transformer (MaxViT), Compact Convolutional Transformers (CCT), Pyramid Vision Transformer (PVT), SLViT and a ViT enabled CNN model called "PlantXViT" work well for high dimensional multicast image datasets. Deep learning-based system evaluation the performance of the models, different metrics such as accuracy, precision, recall, etc. were used in the existing studies. Finally, these technologies are aimed at addressing the challenge of early diagnosis and management of plant diseases.